# Exploring Logic Artificial Chemistries: An Illogical Attempt?

Christof Teuscher
Los Alamos National Laboratory
CCS-3, MS-B265, Los Alamos, NM 87545, USA
christof@teuscher.ch | www.teuscher.ch/christof

*Abstract*— Robustness to a wide variety of negative factors and the ability to self-repair is an inherent and natural characteristic of all life forms on earth. As opposed to nature, man-made systems are in most cases not inherently robust and a significant effort has to be made in order to make them resistant against failures. This can be done in a wide variety of ways and on various system levels. In the field of digital systems, for example, techniques such as triple modular redundancy (TMR) are frequently used, which results in a considerable hardware overhead. Biologically-inspired computing by means of biochemical metaphors offers alternative paradigms, which need to be explored and evaluated.

Here, we are interested to evaluate the potential of nature-inspired artificial chemistries and membrane systems as an alternative information representing and processing paradigm in order to obtain robust and spatially extended Boolean computing systems in a distributed environment. We investigate conceptual approaches inspired by artificial chemistries and membrane systems and compare proof-of-concepts. First, we show, that elementary logical functions can be implemented. Second, we illustrate how they can be made more robust and how they can be assembled to larger-scale systems. Finally, we discuss the implications for and paths to possible genuine implementations. Compared to the main body of work in artificial chemistries, we take a very pragmatic and implementation-oriented approach and are interested in realizing Boolean computations only. The results emphasize that artificial chemistries can be used to implement Boolean logic in a spatially extended and distributed environment and can also be made highly robust, but at a significant price.

## I. Introduction and Motivation

For more than half a century, the von Neumann computer architecture and the abstract Turing machine have largely dominated computer science in many variants and refinements. One might ask, what the future of these two major paradigms will look like. Without disruptive new technologies, it is expected that the ever-increasing computing performance and storage capacity achieved with existing technologies will eventually reach a plateau. To address this challenge, there is a growing interest in novel computing paradigms and machines in order to keep going at the current pace of progress and to face tomorrow's complex large-scale grand challenges.

This quest has been further supported by the appearance of novel materials and fabrication methods, such as nanotechnology and synthetic biology, which have the potential to build large-scale computing systems by a mainly bottom-up (self-)assembled process. Despite important progress in recent years, nanoscale electronics, for example, is still in its infancy and there is no consensus on what type of computing architecture holds most promises. Nevertheless, it is generally assumed that future computing substrates made up from billions or even an Avogadro number (i.e., $6 \times 10^{23}$) of components will be entirely or at least partly irregular, heterogeneous, fine-grained, imperfect, and highly unreliable.

In this paper, we are interested to explore the nature-inspired information representing and processing paradigms of artificial chemistries and membrane systems to realize reliable elementary computing functions in an alternative, distributed, and inherently parallel way, which—so at least the hope—might be useful for emerging computing machines at a later stage. The work is further motivated by the wish to obtain dynamic and "adjustable" robustness by using a minimal amount of resources to take into account changing environments, and to therefore go beyond the traditional and static ways to address robustness, such as triple modular redundancy (TMR) [12] in electronics.

Boolean circuits are the basis of any modern computer, and it would obviously be interesting if we could realize them in an alternative, straightforward, and robust way by means of artificial chemistries. Digital circuits essentially contain elementary logical functions, which are connected together by wires. We therefore have to show that we can implement logical functions, such as AND, OR, NOT, and that we can wire them together in an arbitrary way.

In order to make computing systems robust, both hardware and information redundancy are the main ingredients [12]. Redundancy is something that can more or less straightforwardly be obtained in artificial chemistries because molecules and reactions are easy to maintain in multiple instances, for example.

The critical reader might ask whether realizing Boolean logic on top of a chemical systems is a appropriate and efficient. Whereas alternative ways of representing information and of doing computations in artificial or real chemistries are possible (e.g., see [1]), we are interested in the exploration of alternative building blocks for traditional computers, for which Boolean is the most appropriate, thus the wish to "simulate" Boolean logic.

Although this work is mostly conceptual in its current state, there is a definite effort to make things as realistic and as close to a possible future implementation as possible (see also Section VIII). While most work in the field of membrane computing and artificial chemistries is purely theoretical, our longer-term goal here is in genuine physical realizations by means of alternative computing media. Several candidate substrates are imaginable, e.g., nano- and molecular electronics, bio-engineering, etc.

The remainder of the paper is as following: Section II provides a brief introduction to artificial chemistries and membrane systems. Section III describes the basic framework that we use while Section IV discusses ways to represent information in artificial chemistries. Section V shows several ways to realize Boolean elementary functions and Section VI illustrates a possibility to make them robust by means of redundancy. We can only realize large-scale computing systems if we manage to interconnect and assemble the elementary building blocks. This is shown in Section VII. Finally, Section VIII deals with future implmentational issues and Section IX concludes the paper.

## II. ARTIFICIAL CHEMISTRIES AND MEMBRANE SYSTEMS

Artificial chemistries [8] are man-made systems that are a very general formulation of abstract systems of *objects* that follow arbitrary *rules of interaction*. More formally speaking, an artificial chemistry essentially consist of a set of *molecules* $S$, a set of *rules* $R$, and a definition of the *reactor algorithm* $A$ that describes how the set of rules is applied to the set of molecules. This very broad and appealing paradigm, inspired by bio-chemical systems, allows to describe many complex systems by means of simple rules of decentralized and parallel interactions. Examples are L systems [13], the Gamma language [3], or Fontana's "AlChemy" [9].

In 1998, George Paun initiated *membrane computing* or *P systems* [17]–[19] as an abstract computational model afar inspired by biochemistry and by some of the basic features of biological membranes. A typical membrane system consists of cell-like membranes placed inside a unique "skin" membrane. Multisets of *objects*—usually strings of symbols—and a set of *evolution rules*[1] are then placed inside the regions delimited by the membranes. Each object can be transformed into other objects, can pass through a membrane, or can dissolve or create membranes. The evolution between system configurations is done nondeterministically but synchronously by applying the rules in parallel for all objects able to evolve. With regards to the above definition of artificial chemistries, a membrane system can be considered as such. In this context, a sequence of transitions in a membrane system is called a *computation*. A computation *halts* when a halting configuration is reached, i.e., when no rule can be applied in any region. A computation is considered successful if and only if it halts. However, in the context of intelligent agents, for example, one might imagine systems with a continuous stream of inputs and outputs and no halting configuration.

As an example, let's consider this very simple membrane system, which doesn't do anything useful:

[1]This term is commonly used for membrane systems and artificial chemistries, but *developmental rules* would be more appropriate from a biological point of view.

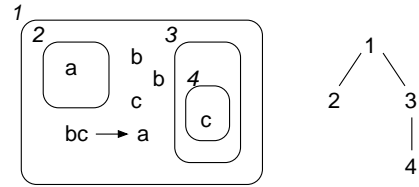

Fig. 1. A four-membrane example membrane system with one cooperative rule in membrane 2. To the left, the rooted-tree structure of the membrane hierarchy is shown. The root corresponds to the skin membrane.

$$\Pi_{example} = (V, \mu, w_1, w_2, w_3, w_4 R_1, R_2, R_3, R_4), \quad (1)$$

where
1) $V = \{a, b, c\}$,
2) $\mu = [_1[_2]_2[_3[_4]_4]_3]_1$,
3) $w_1 = \{a\}$, $w_2 = \{bbc\}$, $w_3 = \{\}$, $w_4 = \{c\}$,
4) $R_1 = R_3 = R_4 = \{\}$, $R_2 = \{bc \to a\}$, .

The membrane structure is depicted in Figure 1, including the associated rooted-tree structure of the membrane hierarchy.

If the membrane chemistry contains rules that have multiple symbols on the left hand side, it is called *cooperative*, otherwise *noncooperative*. The above example is therefore a cooperative chemistry. For more details, the interested reader is referred to [18], [19].

Membrane systems are particularly interesting for the creation of hierarchies, which we consider a key for the creation of complex systems. Hierarchical composition is ubiquitous in physical and biological systems: the nonlinear dynamics at each level of description generates emergent structure, and the nonlinear interactions among these structures provide a basis for the dynamics at the next higher level [21]. But hierarchies are also a means to divide and "hide" complexity, to save resources as the building-blocks might be shared and re-used, and to create higher levels of abstraction.

As we will see in the following, membranes greatly facilitate the implementation of Boolean logical circuits, which are based on elementary building blocks that are assembled together. Among the drawbacks of membrane systems from a rather practical viewpoint is the absence of a design flow (i.e., methodologies and tools) to "program" and develop larger membrane systems and the lack of real "killer-applications."

Finally, note that a wide variety of membrane system flavors exist today. The interested reader is referred to the P systems web site[2] or to [18], [19] for further details.

## III. DESCRIPTION OF THE FRAMEWORK

In order to ready our approach for future large-scale systems, we modify both the formalism and the original membrane systems definition in a few points. The basic idea is to free the system from the need of having any global information transmission and processing capabilities and to open the path for distributed implementations. In a classical membrane system, it is assumed that all rules are applied in each membrane region in a synchronous and maximally parallel manner. While this assumption makes both the development and the analysis of membrane systems easier from an abstract point of view, it potentially raises considerable challenges for physical realizations because global synchronization signals are required. Global (and thus long-distance) connections are costly in terms of resources required (i.e., silicon area) and signal propagation times limit the scalability of the system. To avoid such problems at the conceptual stage already, we use an asynchronous and completely stochastic membrane system, where the reactions within each region are applied in a stochastic manner. We believe that this is also more biologically plausible. Finally, as opposed to classical membrane systems, we also allow the reactions to be rewritten, although we don't directly make use of this feature in this paper.

[2]http://psystems.disco.unimib.it

Formally speaking, the membrane system we use is a construct

$$\Pi = (V, T, \mu, w_1, \ldots, w_m, R_1, \ldots, R_m), \qquad (2)$$

where

1) $V$ is an alphabet. Its elements are called *objects*;
2) $T \subseteq V$ is the *output* alphabet;
3) $\mu$ is a membrane structure consisting of $m$ membranes; $m$ is called the degree of $\Pi$;
4) $w_i, 1 \leq i \leq m$ are strings which represent multisets over $V$ associated with the regions $1, 2, \ldots, m$ of $\mu$;
5) $r$ are evolution rules of the following form:
   - standard reaction $u \to \mathrm{H}v$: replace $u$ by $v$ in the same membrane compartment;
   - output reaction $u \to \mathrm{L}v$: remove $u$ and send $v$ outside the current membrane,
   
   where both $u$ and $v$ can itself be multisets of objects and evolution rules, e.g., $v = ab^2c(a \to b)(b \to c)$. For a better readability, the evolution rules in such an expression shall be put in parenthesis.
6) $R_i, 1 \leq i \leq m$ are finite sets of *evolution rules*.

Note that no complicated rules of the form $u \to v(v_1, in_i)(v_2, out)$ are allowed, which replaces the membrane objects $u$ by the objects $v$, sends the objects $v_1$ to the lower-immediate membrane with label $i$, and the objects $v_2$ to the upper-immediate membrane.

The rules are applied stochastically within each membrane region $i$ as described by the Algorithm 1. Each membrane applies the same algorithm in an asynchronous manner. The motivation behind this reactor algorithm is to open the possibility for distributed multi-reactor implementations, as we'll see in Section VIII. Note that no further parameters, such as the reaction rate, is controlled in this algorithm.

The membrane structure $\mu$ in classical membrane systems always contains a "skin" membrane and the membrane hierarchy is represented by a rooted tree. This sets clear limits on the communication structure among the membranes. In order to circumvent some of these issues, Martín-Vide et al. [16] suggested tissue P systems, which process multisets of

**Algorithm 1** Stochastic reactor algorithm $A$ for a single membrane $i$

1: **while** true **do**
2:     Randomly choose an evolution rule $r = u \to v$ from $R_i$
3:     Randomly choose $|u|$ objects (both molecules and reactions)
4:     **if** $r$ can be applied with these objects **then**
5:         Replace $u$ by $v$ in membrane $i$
6:     **end if**
7: **end while**

symbols in a network of cells, not unlike artificial neural networks. In order to realize any logical function, we not only need to be able to realize elementary logical functions but also be able to interconnect them in an arbitrary way. In this paper, we will use both the rooted-tree and the network-like structure of membrane systems, but without additional finite state memory such as used in tissue P systems.

## IV. INFORMATION AND STATE REPRESENTATION

Probably the biggest factors of success of binary logic in electronics are (1) the ability of signal restoration and (2) the direct correspondence of logical "1" and "0" to voltage or current levels. Although analog computation and machines have existed before digital computers, their success over the years is marginal at best for various reasons. Besides binary and analog systems, multi-valued logic has also found some niche applications in reversible computation. Here, we will focus on binary logic with alternative representations for the logical "1" and "0."

Generally speaking, in order to process information, it has to be represented somehow first, which can be done in a variety of ways. In the field of artificial chemistries, it is natural and straightforward to think of the molecules as an information (or state) carrier and of the reactions as an information processor since they transform molecules into other molecules. The absence or the presence of a single molecule can be interpreted as logical "0" or logical "1" respectively, however, this may cause some trouble when trying to implement a signal inverter. In order to avoid this issue, one can for example represent a logical "1" by the presence of some molecule and a logical "0" by the presence of another one. This is similar to dual-rail logic. However, to avoid ambiguous situations, one has to make sure that at no instant in time, the two molecules are present at the same time. By using multiple molecules, one can also straightforwardly implement multi-valued logic in this way.

The drawback of representing a signal state by a single molecule is the lack of robustness against potential failures: the signal is lost with the loss of the single molecule. In order to make the system robust to certain failures and noise, one can simply represent the signal state in a certain concentration of molecules instead of just a single molecule. With enough redundancy, the signal can then tolerate a certain number of molecule losses. On the other hand, one can also imagine to regenerate the lost molecules and to automatically hold a certain configuration constant. This can be done for example by using the following set of reactions:

$$\begin{aligned} r_1 &= a \to \text{H}a^2 \\ r_2 &= a^m \to \text{H}a^n \end{aligned} \quad (3)$$

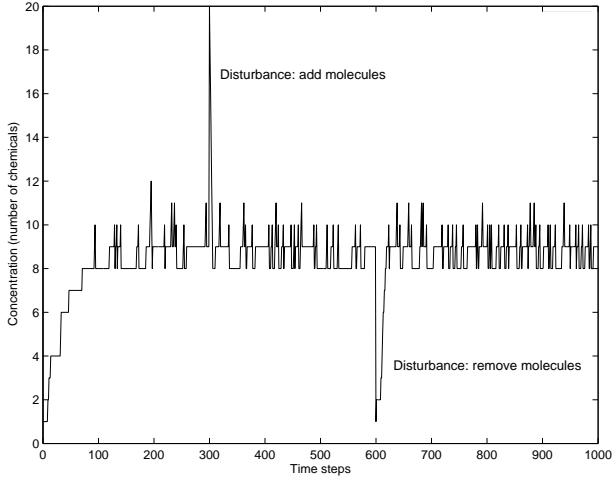

Fig. 2. Constant concentration with disturbances at time steps 300 and 600.

They roughly hold the concentration of $a$ between the upper value $m$ and the lower value $m - n$. Since it is not possible to detect a lower limit for the concentration, we have to constantly generate $a$'s, which will be removed once they reach a concentration of $m$. This results is a saw-tooth-like shape of the concentration. Because of the stochastic nature of the reaction algorithm, the upper and lower limit are not exact. Figure 2 illustrates the behavior of the concentration when started from a single instance of $a$. As one can see, both a negative (remove molecules) and a positive (add molecules) disturbance is quickly corrected. The reaction $r_2$ can also be multiplied to increase the reaction probability and thus the reaction speed.

## V. Elementary Logical Functions

In 2003, Ceterchi and Sburlan [5] presented a solution on how to simulate Boolean circuits with a special class of membrane systems. They first implemented elementary logic functions and then showed that they can be combined to a certain class of more complex circuits, which is limited by the membrane system's inherent tree structure. We think that Ceterchi and Sburlan's implementation is unnecessary complex and illustrate a much simpler solution here[3]. The membrane systems $\Pi_{AND}$ and $\Pi_{NOT}$ realize a AND and NOT function respectively, by using mobile catalysts. Note that we use the classical membrane system notation here, where rules of the form $u \to v_{in} w_{out}$ are allowed, which send $v$ to the inner and $w$ to the outer membrane at the same time. Figure 3 shows the evolution of the $\Pi_{AND}$ membrane system for all possible input combinations.

[3] I am indebted to Petreska Biljana for drawing my attention to this.

$$\Pi_{AND} = (V, C, \mu, w_1, w_2, R_1, R_2), \quad (4)$$

where

1) $V = \{0, 1, d, e, n, x, z, a\}$,
2) $C \subseteq V = \{e, d, a\}$ is the set of mobile catalysts,
3) $\mu = [_1[_2]_2]_1$,
4) $w_1 = \{d, e\}$, $w_2 = \{a, input_1, input_2\}$,
5) $R_1 = \{xa \to a_{in}, e0 \to e_{in}z, ez \to 0_{out}, d1 \to d_{in}, dn \to d1_{out}\}$, $R_2 = \{0a \to 0_{out}a_{out}, 1a \to 1_{out}a_{out}, e0 \to e_{out}x_{out}, e1 \to e_{out}x_{out}, d0 \to d_{out}z_{out}x_{out}, d1 \to d_{out}n_{out}x_{out}\}$.

$$\Pi_{NOT} = (V, C, \mu, w_1, w_2, R_1, R_2), \quad (5)$$

where

1) $V = \{0, 1, d, e, n, x, z\}$,
2) $C \subseteq V = \{n, x\}$ is the set of mobile catalysts,
3) $\mu = [_1[_2]_2]_1$,
4) $w_1 = \{x\}$, $w_2 = \{n, input\}$,
5) $R_1 = \{nx \to n_{in}x\}$, $R_2 = \{0n \to 1_{out}n_{out}, 1n \to 0_{out}n_{out}\}$.

Note that the inputs must be sent in both membrane systems to region 2, which can be a bit tricky to realize. The easiest solution is to add special rules in each compartment that send the input values from the skin membrane to the inner-most membrane.

Since AND and NOT logical functions form a logical basis, i.e., any logical function can be realized from a combination of ANDs and NOTs only, the two above membrane systems could in principle be used to build any logical circuit. However, as mentioned earlier, classical membrane systems only allow for a rooted-tree structure of the membranes, which therefore prevents to build an arbitrary circuit.

Finally, an even simpler—and almost trivial—solution consists in only using cooperative rules without any catalysts. This only requires one membrane and no catalysts have to move back and forth between membranes. Below, only the rules for the four basic logical functions for this single membrane are given:

- $R_{NOT} = \{0 \to \text{L}1, 1 \to \text{L}0\}$
- $R_{AND} = \{00 \to \text{L}0, 01 \to \text{L}0, 11 \to \text{L}1\}$
- $R_{NAND} = \{00 \to \text{L}1, 01 \to \text{L}1, 11 \to \text{L}0\}$
- $R_{OR} = \{00 \to \text{L}0, 01 \to \text{L}1, 11 \to \text{L}1\}$

Note that it is *not* necessary that both input variables arrive at the same time in the membrane as the cooperative rule can only be applied when the two input molecules are available. Once two input symbols are available, the result is directly sent outside (thus the L-symbol in the reactions) the current membrane.

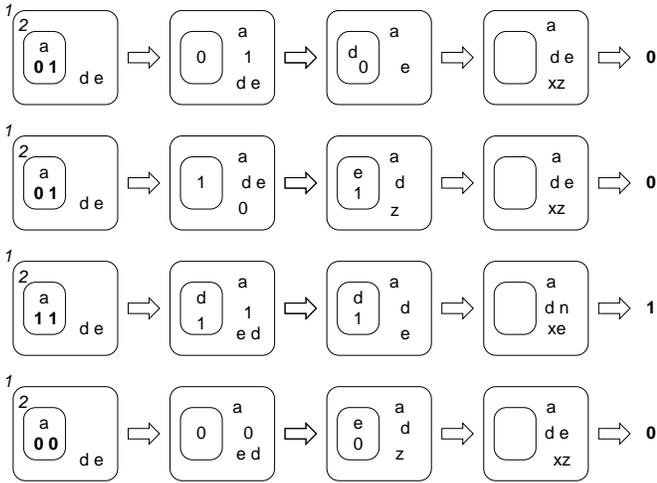

Fig. 3. A self-synchronized AND membrane system with mobile catalysts $e$, $d$, and $a$. The reaction rules are not shown.

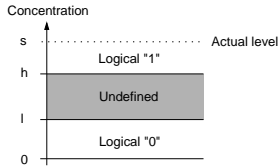

Fig. 4. Interpretation of the chemical concentration of molecules as logical "1" and "0." $s$ indicates the actual concentration of molecules. If $s > h$, we interpret the level as a logical "1," if $s < l$ as logical a logical "0." Intermediate values are undefined.

## VI. ADDING REDUNDANCY

Both hardware and information redundancy (e.g., coding) [12] are the main ingredients for building robust computing systems. As we have seen in Section IV, we can maintain in a fairly straightforward way a concentration of molecules within certain boundaries. Instead of interpreting the presence of a given single molecule as a logical "1," we will consider the presence of a minimal number of molecules as a logical "1." Analogously, if the concentration goes below a certain threshold, we'll consider it as a logical "0." This is not unlike standard CMOS logic, where the voltage levels are also interpreted in a similar way. Figure 4 illustrates the different concentrations and their respective interpretation.

Now, let us extend the simple cooperative chemistry as given at the end of Section V to a more robust chemistry, where the signals are represented by chemical potentials instead of single molecules. The first step consists in modifying the reaction rules of the chemistry as following:

- $R_{NOT} = \{0^h \to L1^m, 1^h \to L0^m\}$
- $R_{AND} = \{0^h0^h \to L0^m, 0^h1^h \to L0^m, 1^h1^h \to L1^m\}$
- $R_{NAND} = \{0^h0^h \to L1^m, 0^h1^h \to L1^m, 1^h1^h \to L0^m\}$
- $R_{OR} = \{0^h0^h \to L0^m, 0^h1^h \to L1^m, 1^h1^h \to L1^m\}$

With regards to the levels as illustrated in Figure 4, $m$ needs to be bigger than $h$, otherwise the chemistry would not correctly interpret the signal levels. The bigger issue we have to address is the following: once enough molecules, i.e., $> h$, are available, the rules will be applied and $m$ molecules as an output will be generated. However, since we want the chemistry to be robust, we initially have to have a larger number than $h$ molecules available in order to guarantee a correct signal level. Let's assume that there are $s > h$ molecules of a certain type within a given membrane. If we remove $h$ molecules by applying one of the above rules, there are $s-h$ molecules remaining, which are superfluous and need to be eliminated, otherwise they will be accumulated over time and will sooner or later be wrongly interpreted as a logical "1." As already stated above, we cannot detect whether a minimal number of molecules is present or not within our membrane system framework. We therefore have to use other "tricks," similar to what we have presented in Section IV to keep a chemical concentration constant. In order to eliminate the unused molecules over time, we simply suggest to introduce the two following rules:

- $r_{0del} = 0^2 \to 0$
- $r_{1del} = 1^2 \to 1$.

Whenever there are more then one 0 or 1 present, these rule will slowly reduce the number towards 0. Since $s >> h$ in general, this reduction will not affect the normal operation and application of the other rules, which need at least $h$ molecules. The only condition one has to satisfy is that there is enough time between two logical operations to allow the remaining molecules to be removed. If that is not guaranteed, one can for example accelerate the molecule reduction by multiplying the above two rules. Similarly, one can also accelerate the information processing by multiplying the logic operation rules, which then increases their probability to be applied successfully.

In this rather naive fault model, which essentially only accounts for lost molecules, we have not dealt with the possibility that one symbol might be accidentally transformed into another one.

In summary: we now have a means to build elementary logical functions with almost arbitrary robustness to "noise" by simply increasing the multiplicity of the molecules which represent a logical value. If molecules get lost due to failures, we can simply increase the tolerance margin for interpreting the concentration and thus—even dynamically, at least in principle—adapt the system to the environment. Obviously this has a cost in terms of the number of symbols one has to apply and eventually store somewhere.

## VII. ASSEMBLING THE ELEMENTARY BUILDING BLOCKS

To show that any logical function can be computed with a given system, one needs to show that the elementary building blocks (e.g., AND, OR, NOT, ...) can be assembled in an arbitrary way. Let us look at an example: in order to compute the Boolean function as shown in Figure 5, a membrane system as illustrated in Figure 6 might be used. The membrane structure is very similar to the structures proposed in [5], but without making use of catalysts. As already stated above, one of the remaining challenges consists in sending the input values to the correct membranes for the initialization. This might for example be realized by means of special rules which only transport the objects from the skin membrane to the inner membranes. In order to avoid that a computation starts when the objects are moved to the inner membranes, they can for example be named differently (e.g., $a$ and $b$ instead of 0 and 1) and be re-transformed to their original symbol at the final destination. Note that the membrane system solution is completely self-timed and synchronized, i.e., a computation starts as soon as the input objects are present.

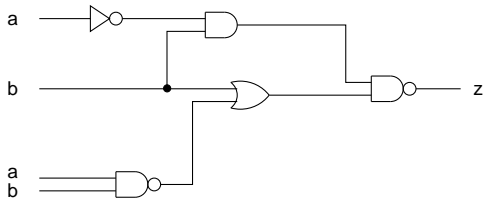

Fig. 5. Example of a logical function with two inputs, $a$ and $b$, and one output $z$. The circuit can be represented as a rooted tree.

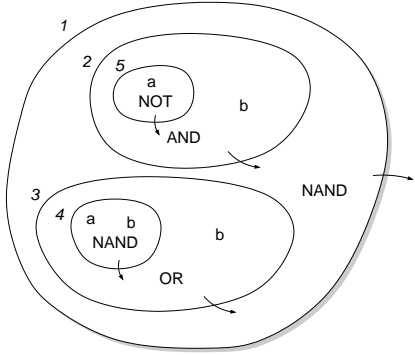

Fig. 6. A membrane system which implements the boolean function as shown in Figure 5. $a$ and $b$ represent the binary input values of the circuit.

Sometimes, the problem arises that one doesn't know when a result is available. This can be addressed by passing a special token along with the computation, so one knows when the token arrives, that the result is ready as well. Another difficulty in larger systems is the occasional need to delay certain results because they depend on other intermediate results before they can proceed. How delays can be realized has been addressed in [26] and shall not be further elaborated here.

As we have seen above, classical membrane systems only allow for membrane hierarchies that can be represented in the form of a rooted tree (see Figure 1). This implies that no recurrent connections are possible in Boolean circuits, which therefore restricts the design space and doesn't allow to implement certain classes of circuits. An alternative representation are membrane assemblies inspired by tissue P systems [16], which process multisets of symbols in a network of cells, not unlike artificial neural networks. Figure 7 illustrates such a network of cells. For more details on how to formalize a tissue P system, the interested reader is referred to [16]. Note that here, we only draw inspiration from the more flexible interconnect topology and not by the other tissue P system particularities.

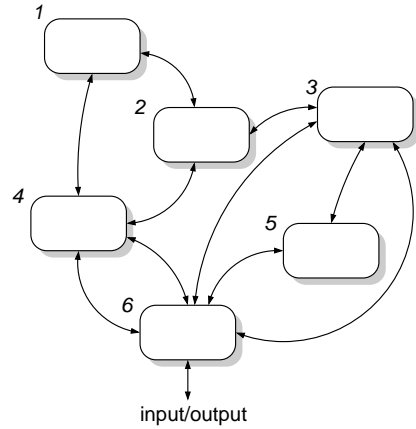

Fig. 7. A network-like assembly of cells, not unlike the tissue P systems [16], allows to build arbitrary interconnect topologies, which is necessary to realize arbitrary logical functions.

If a rule within a cell is applied with the leave-instruction (L) and several outgoing connections to neighboring cells exist (e.g., cell one in Figure 7), then one of the possible connections is chosen at random. Also, if one needs a skin membrane around the network-like assembly of cells as shown in Figure 7, for example to gather output values, that can straightforwardly be added. In fact, a combination of network- and rooted-tree-like structures is in many cases appropriate and useful.

## VIII. IMPLEMENTATIONAL ISSUES

The goal of this section is to reflect about some relevant issues that need to be address if one wants to build a real-world implementation of such an artificial chemistry in a spatially extended way, which might come into reach with future and emerging technologies.

In most computational models driven by abstract considerations, the notions of "space," "size," and "resources" is neglected as there is no need for it. For example, it is irrelevant in a cellular automata model or in a neural network what the dimension of a cell or neuron is. It only matters how it is interconnected with its neighbors and what its functionality is. Likewise, the cells of a membrane system have no particular size attributed and they possess a potentially unlimited capacity when it comes to host symbols and reactions, which is of course not biologically plausible. However, considering space in computational models can be beneficial because "computing in the space domain" is often more attractive than "computing in the time domain," it makes the models more realistic, and it is very helpful when it comes to genuine implementations, where space has to be considered and limited resources are a matter of fact.

As Paun states in [18, p. 367], "[...] membrane computing was a theoretical computer science enterprise, aiming to provide new computational paradigms [...]." The question of how much "added value" (e.g., speed, resources, etc.) a genuine hardware implementation yields with regards to a simulation might—and should—of course always be asked. The most important issue not to forget is, however, that no simulation can be done without real hardware. Thus, the question can basically be reformulated as: "What hardware is appropriate for which simulation?" Although implementing a membrane systems on a traditional sequential von Neumann computer or a distributed arrangement of (usually also sequential) machines is straightforward [6], [7], [22], [23], it remains a simulation as there is not usually parallel hardware involved that could be fully exploited by the inherently parallel membrane systems and their artificial chemistries. As Paun writes, "[i]t is important to underline the fact that "implementing" a membrane system on an existing electronic computer cannot be a real implementation, it is merely a simulation. As long as we do not have genuinely parallel hardware on which the parallelism [...] of membrane systems could be realized, what we obtain cannot be more than simulations, thus losing the main, good features of membrane systems" [18, p. 379]. We believe that high-level simulations of membrane systems is a perfectly valid proof of concept, but that the real challenge of "going back to reality" (see also [18, Chapter 9]) consists in building genuine hardware that is optimized for artificial chemistries and membrane systems. In 2004 [20], Petreska and Teuscher proposed a very first hardware realization of membrane systems using reconfigurable circuits, namely *Field Programmable Gate Arrays* (FPGAs). The implementation is based on a universal and minimal membrane hardware component that allows to very efficiently evolve membrane systems in hardware. Later, Teuscher [24] explored an unconventional and conceptual architecture involving membranes, which was inspired by amorphous computers and randomly interconnected substrates. The architecture remains conceptual, though, and a genuine hardware implementation has yet to be realized.

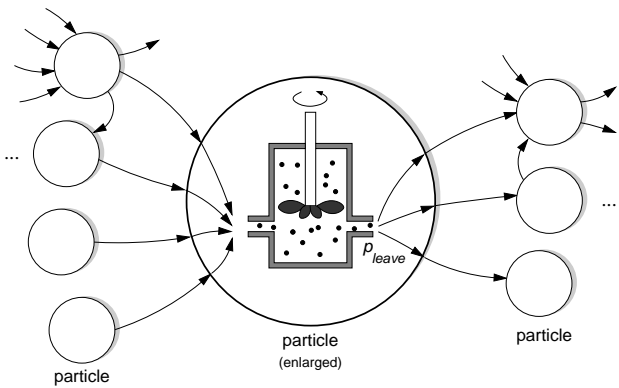

Fig. 8. Simplified view of a particle and its internal stochastic reactor. Each object leaves with a certain probability and can move to a neighboring reactor. The set of reactors can be considered as a large distributed reactor.

From a bird's eye view, the main idea is to implement the cells and membranes on a fine-grained computational substrate of particles, which supports massive parallelism and allows to efficiently run the artificial chemistries. Each particle would contain as a main part an instance of a stochastic reactor. Linked together, the set of reactors forms a large distributed and well-stirred reactor, which is also robust against particle failures, given the system is well enough interconnected and has enough resources available. Figure 8 illustrates the basic idea of this distributed reactor network. In principle, this can be done on a regular, cellular automata like arrangement of cells, however, it is argued that a randomly arranged and interconnected set of particles could more easily be self-assembled by emerging fabrication technologies and materials, such as self-assembling nanowires or nanotubes [25], [27]. In order to implement membranes, which separate sets of reactors from other sets and thus form cells, two main possibilities exist (see Figure 9): possibility (1) seems more natural and simplifies communications, but when an inner cell has to be enlarged, the outer cells have to follow and make room, which can be tricky to implement.

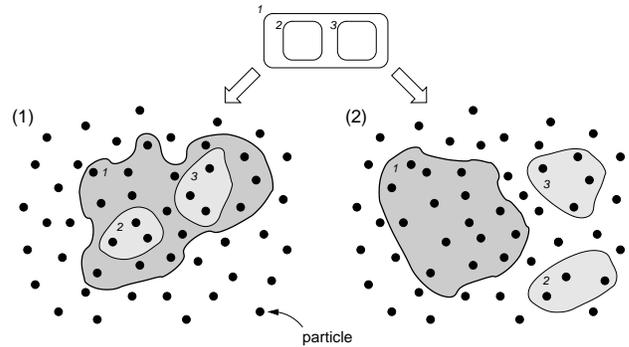

Fig. 9. Two possibilities of implementing hierarchical membrane systems on a programmable reactor multitude.

Many issues need to be explored and resolved, but the goal of this section was not to offer turnkey solutions but rather to raise the awareness of these issues, which will need to be addressed in the future, and where artificial chemistries offer appealing concepts.

## IX. CONCLUSION

In this paper, we have explored a few possibilities of how to realize Boolean logic by means of artificial chemistries, in particular a variant of membrane systems. Compared to others [4], [14]—who also had different goals—we have chosen the path of design-by-hand instead of design by means of evolutionary algorithms. We have shown that elementary Boolean functions can be implemented in various ways and with various complexity. It is also more or less easily possible to make the functions robust my adding redundancy, which can be done in a very scalable way. Next, we have shown that the elementary functions can be assembled into larger systems. Finally, we've also discussed some implementational issues for possible future genuine hardware realizations.

Because of the limited scope, we could not address a number of issues in this work, such as the construction of state machines, clocks (oscillators), synchronous systems, etc. However, current work could readily be extended to build more complex machines, such as state machines, and not just combinational Boolean logic. For example, the method for holding an artificial chemical concentration constant can be used as a building block to implement various other systems, such as chemical state machines. In [11], Hjelmfelt et al. developed chemically based clocked finite-state machines, including decoders, binary adders, and stack memory. The state machines are based on a chemical neural network with clocking mechanisms. The state of the system is modeled by chemical concentrations. The same method can be easily used to construct a universal Turing machine [10]. None of these two papers does, however, deal with the question what logical operations might be performed, given a certain chemical reaction mechanism. In 1997, Magnasco [15] explicitly showed how to construct logic gates and how arbitrary large circuits can be built from such building blocks. In particular, he also illustrated how the output of a logic gate can be fed into an arbitrary number of inputs of other gates without degrading the logic signal. His approach, however, involves no membranes and more realistic chemical kinematics than used here.

In summary, we have seen that implementing Boolean logic by means of membrane systems is feasible in several ways, and can even be very elegant, but that the price to pay (i.e., the overhead) can be significant, especially if the system needs to be robust against failures and noise. Furthermore, the membranes of membrane systems provide a unique feature that allows to nicely build compartments and to use them as building blocks in order to realize more complex systems. This can be well exploited by how we build logical systems (i.e., "assembling blocks to form new blocks, ...").

Whether the presented approach will be competitive with any other conventional or unconventional realization of Boolean logic in the future, remains an open question and further investigation and especially comparisons are necessary. It is to be expected that signal propagation times, for example, will play at least an equally important roles as in classical systems, and we might thus well reach the same scalability limitations at some point.

Future work will concentrate on pursuing a more systematic exploration of the design trade-offs and to compare larger-scale examples in terms of the number of resources used, communication cost and delays, and other relevant factors.